\documentclass[10pt, conference, compsocconf]{IEEEtran}
\usepackage[pdftex]{graphicx}
\usepackage[pagebackref=true,breaklinks=true,colorlinks,bookmarks=false]{hyperref}
\usepackage{times}
\usepackage{epsfig}
\usepackage{graphicx}
\usepackage{amsmath}
\usepackage{amssymb}
\usepackage{subfig}
\usepackage{bbm}
\usepackage[belowskip=-15pt,aboveskip=0pt]{caption}
\newcommand{\fig}[1]{Fig.~\ref{#1}} 
\usepackage{tabularx}
\usepackage{array}
\usepackage{multirow}
\usepackage{booktabs}
\usepackage{authblk}
\usepackage{color, colortbl}
\definecolor{Gray}{gray}{0.9}
\begin{document}
\title{In Defense of Classical Image Processing: Fast Depth Completion on the CPU}

\author[]{ Jason Ku}
\author[]{ Ali Harakeh}
\author[]{ Steven L. Waslander}
\affil[] {
Mechanical and Mechatronics Engineering Department
\\
University Of Waterloo
\\
Waterloo, ON, Canada
}
\affil[]{\textit{jason.ku@uwaterloo.ca, www.aharakeh.com, stevenw@uwaterloo.ca}}

\maketitle

\begin{abstract}
With the rise of data driven deep neural networks as a realization of universal function approximators, most research on computer vision problems has moved away from hand crafted classical image processing algorithms. This paper shows that with a well designed algorithm, we are capable of outperforming neural network based methods on the task of depth completion. The proposed algorithm is simple and fast, runs on the CPU, and relies only on basic image processing operations to perform depth completion of sparse LIDAR depth data. We evaluate our algorithm on the challenging KITTI depth completion benchmark \cite{geiger2012we}, and at the time of submission, our method ranks $first$ on the KITTI test server among all published methods. Furthermore, our algorithm is data independent, requiring no training data to perform the task at hand. The code written in Python will be made publicly available at \href{https://github.com/kujason/ip\_basic}{https://github.com/kujason/ip\_basic}.
\end{abstract}

\begin{IEEEkeywords}
image processing; depth completion.
\end{IEEEkeywords}

\IEEEpeerreviewmaketitle

\section{Introduction}
The realization of universal function approximators via deep neural networks has revolutionized computer vision and image processing. Deep neural networks have been used to approximate difficult high dimensional functions involved in object detection \cite{ku2017joint}, semantic and instance level segmentation \cite{zhao2016pyramid}, and even the decision making process for driving a car \cite{bojarski2016end}. The success of these function approximators on AI-complete \cite{yampolskiy2012ai} tasks has lead the research community to stray away from classical non-learning based methods to solve almost all problems. This paper aims to show that well-designed classical image processing algorithms can still provide very competitive results compared to deep learning based methods. We specifically tackle the problem of depth completion, that is, inferring a \textbf{dense} depth map from image and \textbf{sparse} depth map inputs.

Depth completion is an important task for machine vision and robotics. Current state-of-the-art LIDAR sensors can only provide sparse depth maps when projected back to image space. This limits both the performance and the operational range of many perception algorithms that rely on the depth as input. For example, 3D object detection algorithms \cite{ku2017joint,cvpr17chen,qi2017frustum} can regress bounding boxes only if there are enough points belonging to the object.

Many different approaches have been proposed for depth completion. These approaches range from simple bilateral upsampling based algorithms \cite{chan2008noise} to end-to-end deep learning based ones \cite{uhrig2017sparsity}. The latter are very attractive as they require minimal human design decisions due to their data driven nature. However, using deep learning approaches results in multiple consequences. First, there is finite compute power on embedded systems. GPUs are very power hungry, and deploying a GPU for each module to run is prohibitive. Second, the creation of deep learning models without proper understanding of the problem can lead to sub-optimal network designs. In fact, we believe that solving this problem with high capacity models can only provide good results after developing sufficient understanding of its underlying intricacies through trying to solve it with classical image processing methods.

\begin{figure*}[t]
\begin{center}
\includegraphics[width=\textwidth]{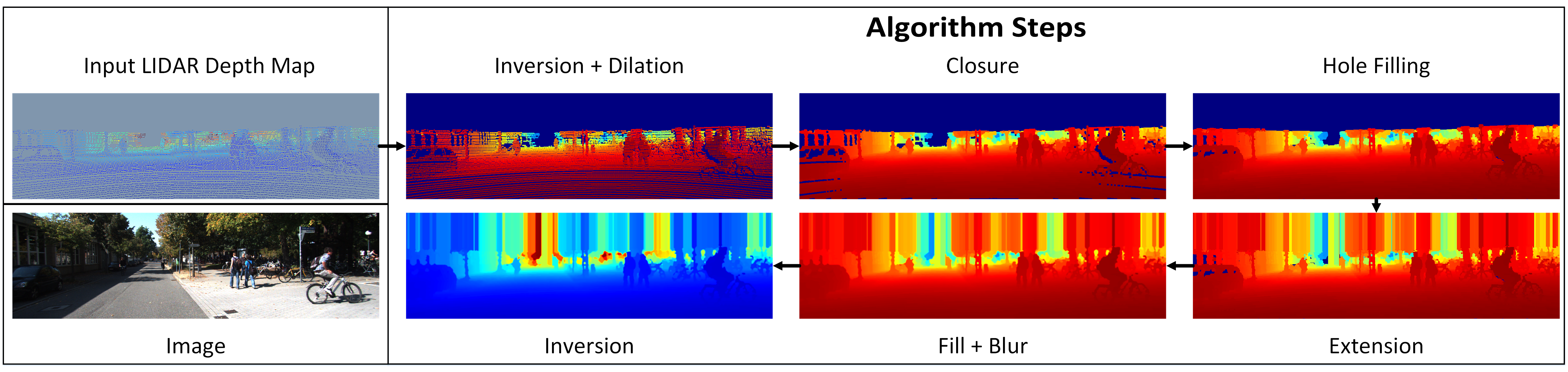}
\caption{A flowchart of the proposed algorithm. \textbf{Clockwise starting at top left:} Input LIDAR depth map (enhanced for visibility), inversion and dilation, small hole closure, small hole fill, extension to top of frame, large hole fill and blur, inversion for output, image of scene (not used, only for reference).}
\label{algorithm}
\end{center}
\end{figure*}

This paper aims to show that on certain problems, deep learning based approaches can still be outperformed by well designed classical image processing based algorithms. To validate this, we design a simple algorithm for depth completion that relies on image processing operations only. The algorithm is non-guided and relies on LIDAR data only, making it independent of changes in image quality. Furthermore, our algorithm is not deep learning based, requiring no training data, making it robust against overfitting. The algorithm runs as fast as deep learning based approaches but on the CPU, while performing better than the custom designed sparsity invariant convolutional neural network of \cite{uhrig2017sparsity}. To summarize, our contributions are as follows:
\begin{itemize}
\item We provide a fast depth completion algorithm that runs at $90$ Hz on the CPU and ranks \textit{first} among all published methods on the KITTI depth completion benchmark \cite{kitti}.
\item We show that our algorithm outperforms CNN based approaches that have been designed to tackle sparse input representations by a wide margin.
\end{itemize}

The rest of this paper is structured as follows: Section \ref{rwork} provides a brief overview of state-of-the-art depth completion algorithms. Section \ref{alg} describes the problem of depth completion from a mathematical perspective and then introduces our proposed algorithm. Section \ref{exp} provides a qualitative and quantitative comparison with the state-of-the-art methods on the KITTI depth completion benchmark. Finally, we conclude the paper with Section \ref{conc}.
\section{Related Work}
\label{rwork}
Depth completion or upsampling is an active area of research with applications in stereo vision, optical flow, and 3D reconstruction from sparse LIDAR data. This section discusses state-of-the-art depth completion algorithms while categorizing them into two main classes: guided depth completion and non-guided depth completion.\\

\noindent\textbf{Guided Depth Completion:} Methods belonging to this category rely on colour images for guidance to perform depth map completion. A variety of previous algorithms have proposed joint bilateral filtering to perform ``hole filling" on the target depth map \cite{qi2013structure,chen2012depth,richardt2012coherent}. Median filters have also been extended to perform depth completion from colour image guidance \cite{matyunin2011temporal}. Recently, deep learning approaches have emerged to tackle the guided depth completion problem \cite{hui2016depth, song2016deep}. These methods have been demonstrated to produce higher quality depth maps, but are data-driven, requiring large amounts of training data to generalize well. Furthermore, these algorithms assume operation on a regular grid and fail when applied to very sparse input such as the depth map output of a LIDAR sensor.
All of the above guided depth completion models suffer from a dependency on the quality of the guiding colour images. The performance on the depth completion task deteriorates as the quality of the associated colour image becomes worse. Furthermore, the quality of the depth map output is highly correlated to the quality of calibration and synchronization between the depth sensor and the camera. Our proposed algorithm is non-guided and requires no training data, resolving most of the problems faced with the guided depth completion approach.\\

\noindent\textbf{Non-Guided Depth Completion:} Methods belonging to this category use only a sparse depth map to produce a dense one. \cite{hornacek2013depth} uses repetitive structures to identify similar patches in 3D across different scales to perform depth completion. \cite{uhrig2017sparsity} provides a baseline using Nadaraya-Watson kernel regression \cite{nadaraya1964estimating} to estimate missing values for depth completion of sparse LIDAR scans. Missing values do not contribute to the Gaussian filter and therefore sparsity is implicitly handled in the algorithm. \cite{uhrig2017sparsity} recently proposed a sparsity invariant CNN architecture for depth completion. The proposed sparsity invariant convolutional layer only considers ``valid" values in the output computation providing better results than normal convolutional kernels. However, \cite{uhrig2017sparsity} also provided results for CNNs trained on nearest neighbour interpolated depth maps that outperformed these sparsity invariant CNNs, diminishing the practical value of pursuing this direction of research for depth completion. As discussed in the previous sections, deep learning based approaches are still too computationally taxing, requiring systems to deploy power hungry GPUs to run instances of the neural network. In the next sections, we aim to show that our classical image processing algorithm can perform as well as deep neural networks and at a similar frame rate without incurring additional restrictions on the deployment hardware.     

\begin{figure}[b]
\begin{center}
\includegraphics[width=\columnwidth]{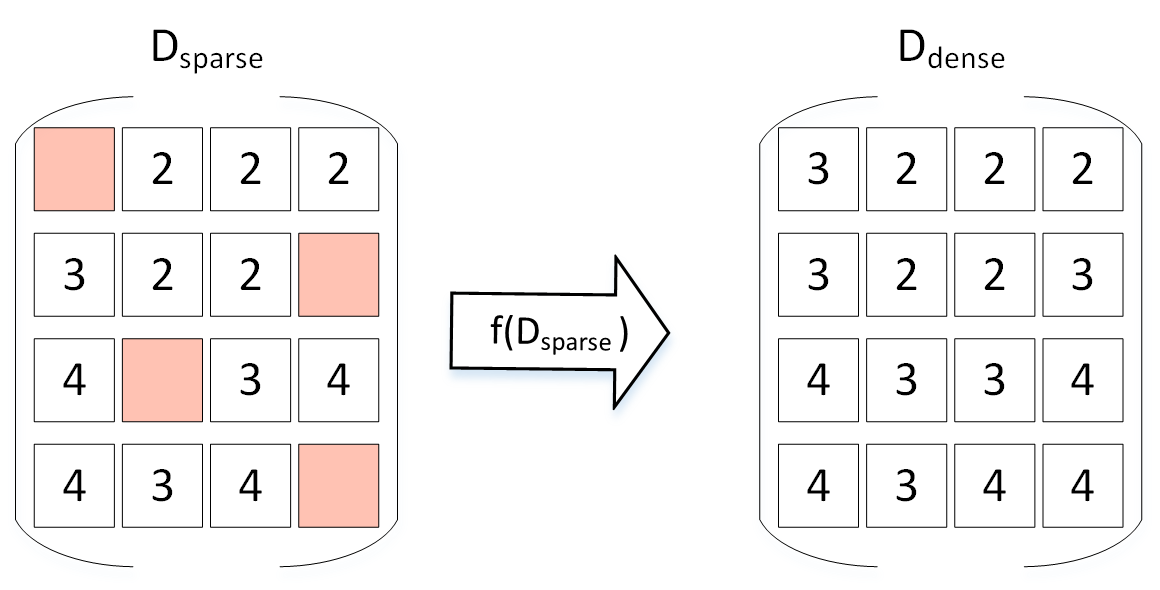}
\caption{A toy example summarizing the problem formulation described in equation \ref{prob_formul}. Empty values are coloured in red, and filled by applying the function $f$ to $D_{sparse}$.}
\label{prob_form}
\end{center}
\end{figure}

\section{Problem Formulation}
The problem of depth completion can be described as follows: 

\noindent Given an image $I \in \mathbb{R}^{M \times N}$, and a \textit{sparse} depth map $D_{sparse} \in \mathbb{R}^{M \times N}$ find $\hat{f}$ that approximates a true function $f: \ \mathbb{R}^{M \times N} \times \mathbb{R}^{M \times N} \rightarrow \mathbb{R}^{M \times N}$ where $f(I,D_{sparse}) = D_{dense}$. The problem can be formulated as: 
\begin{equation}
\begin{aligned}
min. \ ||\hat{f}(I,D_{sparse}) - f(I,D_{sparse})||_F^2 = 0 
\end{aligned}
\label{prob_formul}
\end{equation} 
Here, $D_{dense}$ is the output dense depth map, and has the same size as $I$ and $D_{sparse}$ with empty values replaced by their depth estimate. In the case of non-guided depth completion, the above formulation becomes independent of the image $I$ as shown in Fig. \ref{prob_form}. We realize $\hat{f}$ via a series of image processing operations described below. 

\section{Proposed Algorithm}
\label{alg}
The proposed method, as shown in \fig{algorithm} is implemented in Python and uses a series of OpenCV \cite{opencv} and NumPy \cite{numpy} operations to perform depth completion. We leverage the implementation of standard OpenCV operations, which use larger pixel values to overwrite lower pixel values. This way, the issue of sparsity can be addressed by selecting appropriate operations to fill in empty pixels. By exploiting this property of OpenCV operations, we realize depth completion via the eight step algorithm described below.

\begin{figure}[t]
    \centering
    \includegraphics[width=0.7\columnwidth]{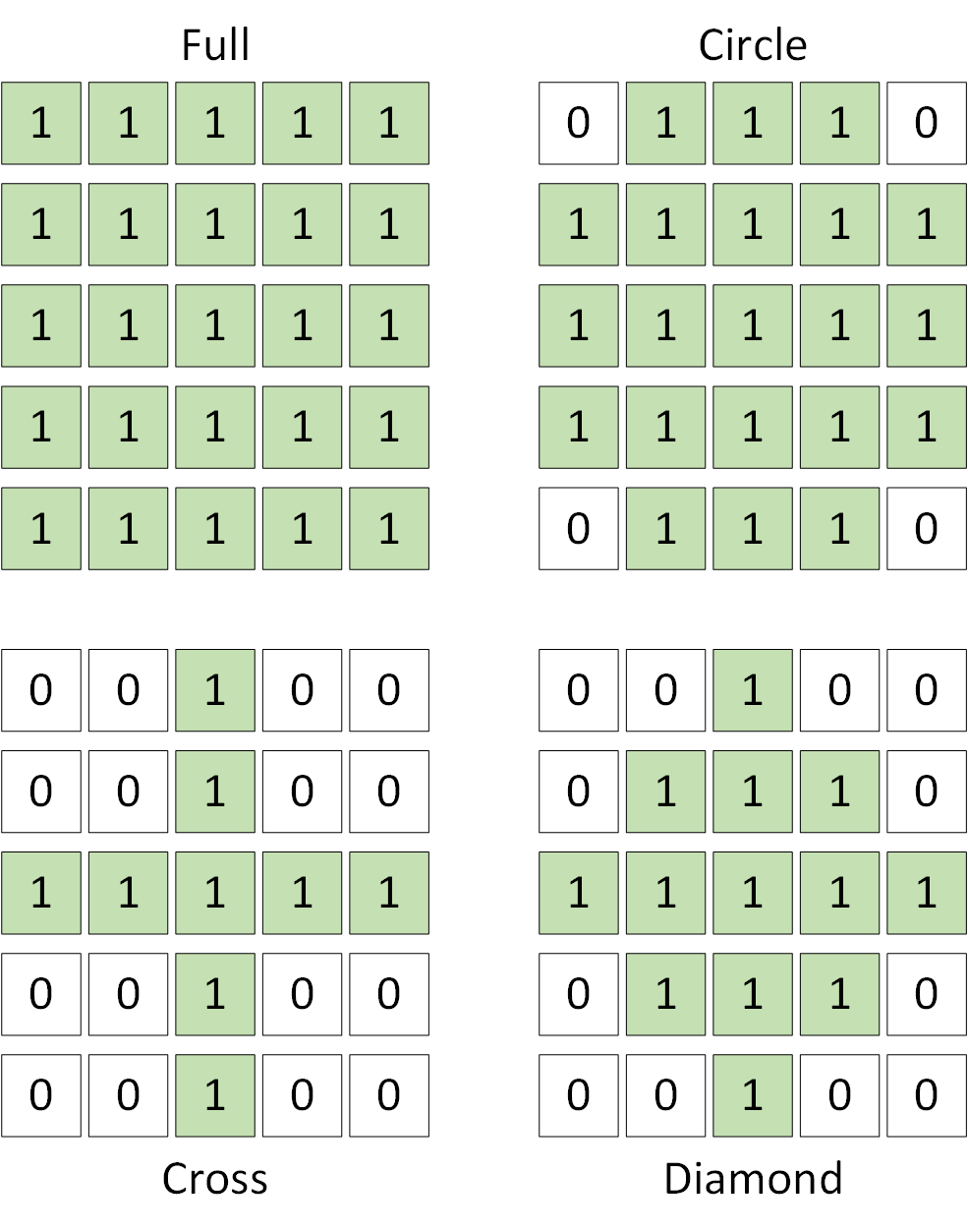}
    \caption{Different kernels used for comparison.}
    \label{kernels}
\end{figure}

The final result of the algorithm is a dense depth map $D_{dense}$ that can be used as input for 3D object detection, occupancy grid generation, and even simultaneous localization and mapping (SLAM).

\subsubsection{Depth Inversion}
The main sparsity handling mechanisms employed are OpenCV morphological transformation operations, which overwrite smaller pixel values with larger ones. When considering the raw KITTI depth map data, closer pixels take values close to $0$ m while further ones take values up to a maximum of $80$ m. However, empty pixels take the value $0$ m too, which prevents using native OpenCV operations without modification. Applying a dilation operation on the original depth map would result in larger distances overwriting smaller distances, resulting in the loss of edge information for closer objects. To resolve this problem, valid (non-empty) pixel depths are inverted according to $D_{inverted} = 100.0 - D_{input}$, which also creates a $20$ m buffer between valid and empty pixel values. This inversion allows the algorithm to preserve closer edges when applying dilation operations. The $20$ m buffer is used to offset the valid depths in order to allow the masking of invalid pixels during subsequent operations.

\subsubsection{Custom Kernel Dilation}
We start by filling empty pixels nearest to valid pixels, as these are most likely to share close depth values with valid depths. Considering both the sparsity of projected points and the structure of the LIDAR scan lines, we design a custom kernel for an initial dilation of each valid depth pixel. The kernel shape is designed such that the most likely pixels with the same values are dilated to the same value. We implement and evaluate four kernel shapes shown in \fig{kernels}. From the results of the experiments performed in Section \ref{exp}, a $5 \times 5$ \textit{diamond} kernel is used to dilate all valid pixels.

\subsubsection{Small Hole Closure}
After the initial dilation step, many holes still exist in the depth map. Since these areas contain no depth values, we consider the structure of objects in the environment and note that nearby patches of dilated depths can be connected to form the edges of objects. A morphological close operation, with a $5 \times 5$ \textit{full} kernel, is used to close small holes in the depth map. This operation uses a binary kernel, which preserves object edges. This step acts to connect nearby depth values, and can be seen as a set of $5 \times 5$ pixel planes stacked from farthest to nearest.

\begin{figure*}[t]
\begin{center}
\includegraphics[width=\textwidth]{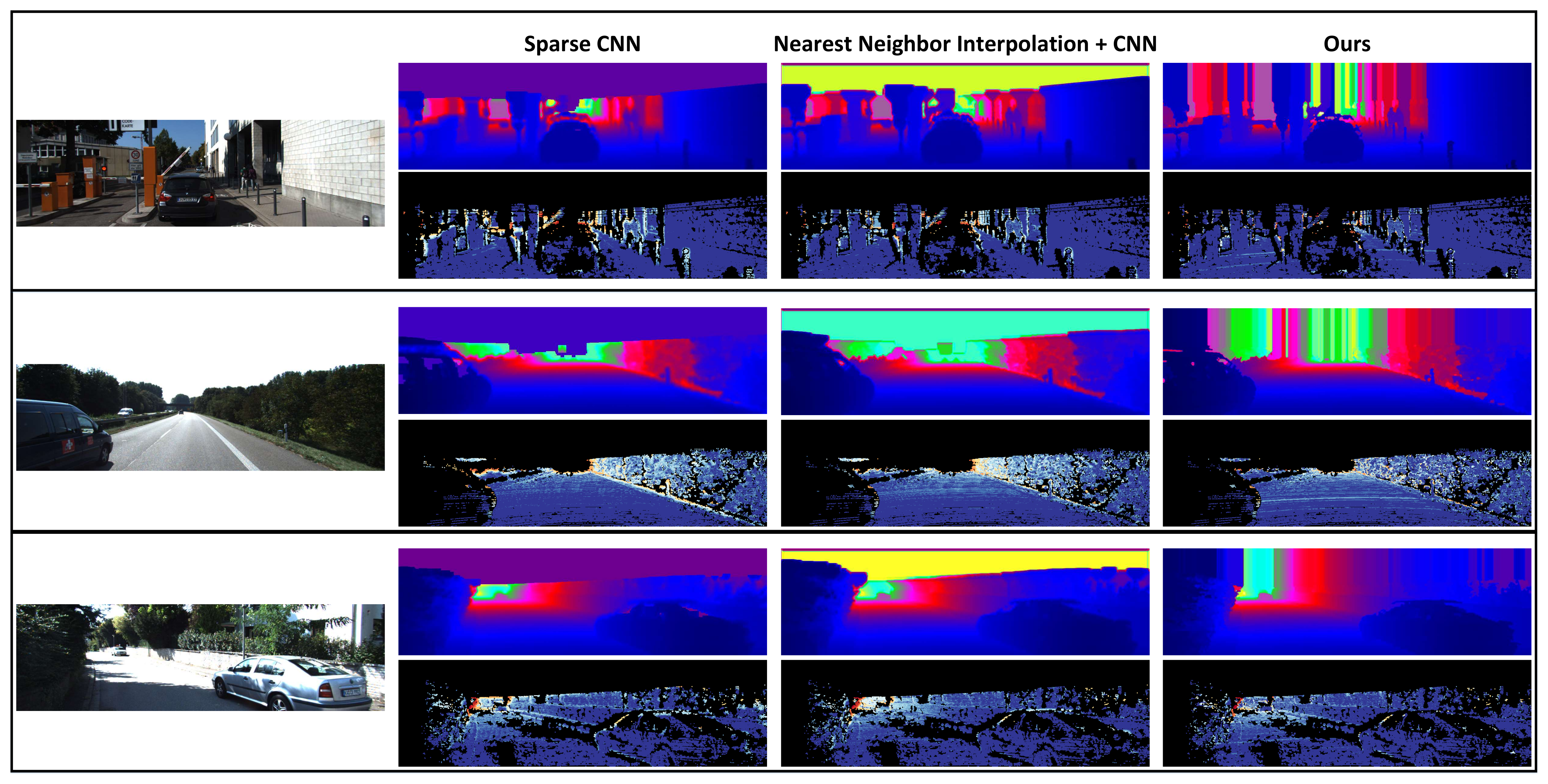}
\caption{The qualitative results of our proposed algorithm on three samples in the KITTI test set in comparison to Sparse CNN and Nearest Neighbour Interpolation with CNN, both of which were proposed in \cite{uhrig2017sparsity}. \textbf{Top}: Output dense depth map. \textbf{Bottom:} Visualization of the pixel-wise error in estimation ranging from \textbf{blue} for a low error to \textbf{red} for a high error. It can be seen that our method has a lower error in estimation especially for further away pixels.}
\label{res}
\end{center}
\end{figure*}

\subsubsection{Small Hole Fill}
Some small to medium sized holes in the depth map are not filled by the first two dilation operations. To fill these holes, a mask of empty pixels is first calculated, followed by a $7 \times 7$ \text{full} kernel dilation operation. This operation results in only the empty pixels being filled, while keeping valid pixels that have been previously computed unchanged.

\subsubsection{Extension to Top of Frame}
To account for tall objects such as trees, poles, and buildings that extend above the top of LIDAR points, the top value along each column is extrapolated to the top of the image, providing a denser depth map output.

\subsubsection{Large Hole Fill}
The final fill step takes care of larger holes in the depth map that are not fully filled from previous steps. Since these areas contain no points, and no image data is used, the depth values for these pixels are extrapolated from nearby values. A dilation operation with a $31x31$ full kernel is used to fill in any remaining empty pixels, while leaving valid pixels unchanged.

\subsubsection{Median and Gaussian Blur}
After applying the previous steps, we end up with a dense depth map. However, outliers exist in this depth map as a by-product of the dilation operations. To remove these outliers, we use a $5 \times 5$ kernel median blur. This denoising step is very important as it removes outliers while maintaining local edges. Finally, a $5 \times 5$ Gaussian blur is applied in order to smooth local planes and round off sharp object edges.

\subsubsection{Depth Inversion}
The final step of our algorithm is to revert back to the original depth encoding from the inverted depth values used by the previous steps of the algorithm. This is simply calculated as $D_{output} = 100.0 - D_{inverted}$.
\begin{table*}[t]
\centering
\resizebox{0.99\textwidth}{!}{
\begin{tabular}{c|cccc|c}
Method & iRMSE (1/km) & iMAE (1/km) & RMSE (mm) & MAE (mm) & Runtime (s) \\
\midrule
NadarayaW & 6.34 & 1.84 & 1852.60 & 416.77 & 0.05 \\
SparseConvs & 4.94 & 1.78 & 1601.33 & 481.27 & \textbf{0.01} \\
NN+CNN & \textbf{3.25} & \textbf{1.29} & 1419.75 & 416.14 & 0.02 \\
Ours (IP-Basic) & 3.78 & \textbf{1.29} & \textbf{1288.46} & \textbf{302.60} & 0.011 \\ 
\bottomrule
\end{tabular}
}
\caption{A comparison of the performance of Nadaraya-Watson kernel baseline, Sparse CNN, Nearest Neighbour Interpolation with CNN, and our method, evaluated on the KITTI depth completion \textit{test} set. Results are generated by KITTI's evaluation server \cite{kitti}.} 
\label{comparison_val}
\end{table*}

\section{Experiments and Results}
\label{exp}
We test our algorithm's performance on the depth completion task in the KITTI depth completion benchmark. The recently released depth completion benchmark contains a large set of LIDAR scans projected into image coordinates to form depth maps. The LIDAR points are projected to the image coordinates using the front camera calibration matrices, resulting in a sparse depth map with the same size as the RGB image. The sparsity is induced by the fact that LIDAR data has a much lower resolution than the image space it is being projected to. Due to the angles of LIDAR scan lines, only the bottom two-thirds of the depth map contain points. The sparsity of the points in the bottom region of the depth maps is found to range between $5-7\%$. The corresponding RGB image is also provided for each depth map, but is not used by our unguided depth completion algorithm. The provided \textit{validation} set of $1000$ images is used for evaluation for all experiments, and the final results on the $1000$ image \textit{test} set are submitted and evaluated by KITTI's test server. The performance of the algorithm and the baselines are evaluated using the inverse Root Mean Squared Error (iRMSE), inverse Mean Average Error (iMAE), Root Mean Squared Error (RMSE), and Mean Average Error (MAE) metrics. We refer the reader to \cite{uhrig2017sparsity} for a deeper insight on each of these metrics. Since methods are ranked based on RMSE on KITTI's test server, the RMSE metric is used as the criterion for selecting the best design.

\subsection{Performance on the Depth Completion Task}
At the time of submission, the proposed algorithm ranks first among all published methods in both RMSE and MAE metrics. Table \ref{comparison_val} provides the results of comparison against the baseline Nadaraya-Watson kernel method (NadarayaW), as well as the learning based approaches Sparsity Invariant CNNs (SparseConvs) and Nearest Neighbour Interpolation with CNN (NN+CNN) \cite{uhrig2017sparsity}, all of which are specifically tailored for processing sparse input. Our algorithm outperforms the NN+CNN, the runner up on the KITTI data set, by 131.29 mm in RMSE and 113.54 mm in MAE. That is equivalent to a difference of 11 cm mean error in the final point cloud results, which is important for accurate 3D object localization, obstacle avoidance, and SLAM. Furthermore, our proposed algorithm runs at 90 Hz on an Intel® Core™ i7-7700K Processor, while both the second and third ranking methods require an additional GPU to run at 50 and 100 Hz respectively. 

\subsection{Experimental Design}
To design the algorithm, a greedy design procedure is followed. Since empty pixels nearby valid pixels are likely to share similar values, we structure the order of the algorithm with smaller to larger hole fills. This allows the area of effect for each valid pixel to increase slowly while still preserving local structure. The remaining empty areas are then extrapolated, but have become much smaller than before. A final blurring step is used to reduce output noise and smooth out local planes.

The effect of design choices for the dilation kernel sizes are first explored, followed by those of that kernel's shape, and finally the blurring kernels employed after dilation. We choose the best result of each experiment to continue with the next design step. Due to this greedy design approach, the first two experiments on kernel size and shape do not include the blurring of Step 7. The final algorithm design uses the top performing designs from each experiment to achieve the best result. \\

\noindent\textbf{Custom Kernel Design:}
The design of the initial dilation kernel is found to greatly affect the performance of the algorithm. To find an optimal dilation kernel, a \textit{full} kernel is varied between $3\times3$, $5\times5$, and $7\times7$ sizes. A $7\times7$ kernel is found to dilate depth values past their actual area of effect, while a $3\times3$ kernel dilation does not expand pixels enough to allow edges to be connected by later hole closing operations. Table \ref{compare_size} shows that a $5\times5$ kernel provides the lowest RMSE.

Using the results of the kernel size experiment, the design space of $5\times5$ binary kernel shapes is explored. A full kernel is used as a baseline, and compared with circular, cross, and diamond kernel shapes. The shape of the dilation kernel defines the initial area of effect for each pixel. Table \ref{compare_size} shows that a diamond kernel provides the lowest RMSE. The diamond kernel shape preserves the rough outline of rounded edges, while being large enough to allow edges to become connected by the next hole closing operation. It should be noted that the size and shape of the dilation kernel is not found to have a significant impact on runtime.\\

\begin{table}[t]
\centering
\resizebox{0.99\columnwidth}{!}{
    \begin{tabular}{c|cc}
        Kernel Size & RMSE (mm) & MAE (mm)\\
        \midrule
        3x3 & 1649.97 & 367.06  \\
        5x5 & \textbf{1545.85} & \textbf{349.45} \\
        7x7 & 1720.79 & 430.82 \\
        \midrule
         Kernel Shape & RMSE (mm) & MAE (mm) \\
        \midrule
        Full & 1545.85 & 349.45 \\
        Circle & 1528.45 & 342.49 \\
        Cross & 1521.95 & 333.94 \\
        Diamond & \textbf{1512.18} & \textbf{333.67} \\ 
        \bottomrule
    \end{tabular}
}
\caption{Effect of dilation kernel shape and size on the performance of the algorithm. The algorithm design is optimized in a greedy fashion for kernel size first, then for kernel shape.}
\label{compare_size}
\end{table}

\begin{table}[b]
\centering
\resizebox{0.99\columnwidth}{!}{
    \begin{tabular}{c|cc|c}
        Kernel & RMSE (mm) & MAE (mm) & Runtime (s) \\
        \midrule
        No Blur & 1512.18 & 333.67 & 0.007 \\
        Bilateral Blur & 1511.80 & 334.12 & 0.011 \\
        Median Blur & 1461.54 & 323.34 & 0.009 \\
        Median + Bilateral Blur & 1456.69 & 328.02 & 0.014 \\
        Gaussian Blur & 1360.06 & 310.39 & 0.008 \\
        Median + Gaussian Blur & \textbf{1350.93} & \textbf{305.35} & 0.011 \\
        \bottomrule
    \end{tabular}
}
\caption{Effect of blurring.}
\label{compare_blur}
\end{table}

\begin{figure*}[t]
    \centering
    \includegraphics[width=0.99\textwidth]{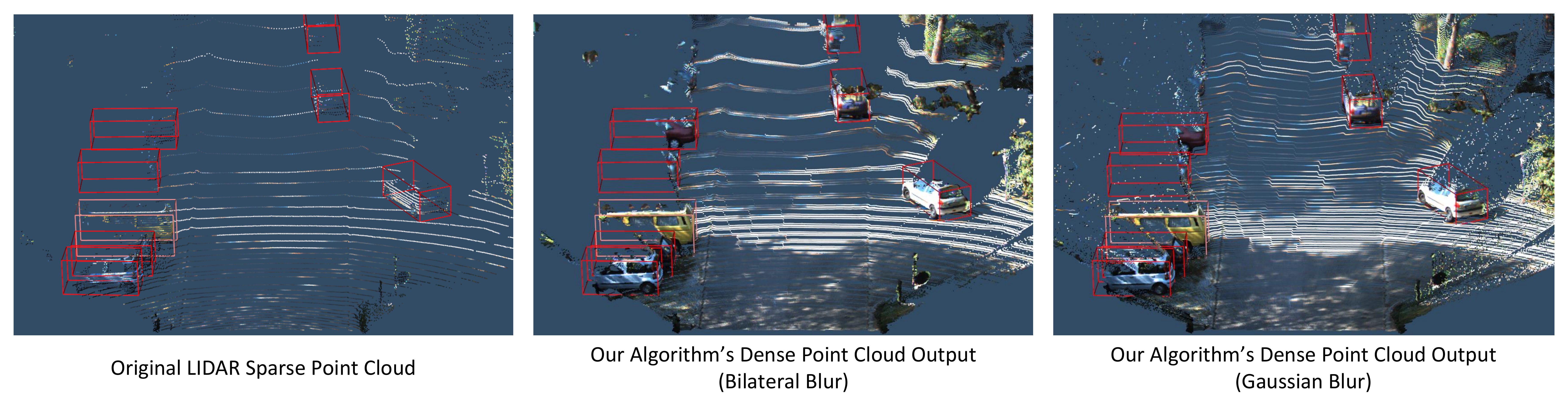}
    \caption{Qualitative results of the application of our algorithm on LIDAR point clouds using both bilateral and Gaussian blur kernels. Points have been colourized using the RGB image for easier visualization. It can be seen that ground truth object detection bounding boxes in \textbf{red} have much more points in the dense point cloud.}
    \label{pointcloud_res}
\end{figure*}

\noindent\textbf{Noise Reduction through Blurring:}
The depth map output contains many small flat planes and sharp edges due to the \textit{Manhattan} \cite{coughlan2001manhattan} nature of the environment, and the series of binary image processing operations applied during the previous steps. Furthermore, small areas of outliers may be dilated, providing erroneous patches of depth values. To apply smoothing to local planes, round off object edges, and remove outlier depth pixels, we study the effect of median, bilateral, and Gaussian blurring on the algorithm's performance.

Table \ref{compare_blur} shows the effect of different blur methods on the final performance of the algorithm and on its runtime. A median blur is designed to remove salt-and-pepper noise, making it effective in removing outlier depth values. This operation adds 2 ms to the runtime, but the improvement can be seen through a decrease in both RMSE and MAE. A bilateral blur preserves local structure by only blurring pixels with other nearby pixels having similar values, and has minimal effect on the evaluated RMSE and MAE metrics, while adding 4 ms to the runtime. Due to the Euclidean calculation of the RMSE metric, a Gaussian blur reduces RMS errors significantly by minimizing the effect of outlier pixel depths. The Gaussian blur also runs fastest, adding only 1 ms to the average runtime. The final algorithm employs a combined median and Gaussian blur as this combination is shown to provide the lowest RMSE.

Figure \ref{pointcloud_res} shows the results of running the algorithm with two different blurring kernels on a projected point cloud from a sample in the KITTI object detection benchmark. The steps of extending depth values to the top of frame and large hole filling are skipped since they introduce a large number of extrapolated depth values. For applications where a fully dense depth map is not required, it is recommended to limit both the upward extension per column and dilation kernel size. While the Gaussian blur version provides the lowest RMSE, it also introduces many additional 3D points to the scene. The bilateral blur version preserves the local structure of objects and is recommended for practical applications. It should be noted that the points are colourized using the RGB image, but image data is not used in our unguided approach. An accurate, denser point cloud can be helpful for 3D object detection methods \cite{ku2017joint,cvpr17chen,zhou2017voxelnet} which rely on point cloud data for both object classification and localization. After depth completion, it can be seen that ground truth labelled objects contain many more points. The structure of the cars and road scene become much clearer, and this is especially noticeable for the objects farther away. More qualitative results are available in video format at \href{https://youtu.be/t\_CGGUE2kEM}{https://youtu.be/t\_CGGUE2kEM}.

\section{Conclusion}
\label{conc}
In this work, we propose a depth completion algorithm that takes as an input a sparse depth map to output a dense depth map. Our proposed algorithm uses only traditional image processing techniques and requires no training, making it robust to overfitting. We show that our image processing based algorithm provides state of the art results on the KITTI depth completion benchmark, outperforming several learning-based methods. Our algorithm also runs in real time at 90 Hz and does not require any additional GPU hardware, making it a competitive candidate to be deployed on embedded systems as a preprocessing step for more complex tasks such as SLAM or 3D object detection. Finally, this work is not meant to undermine the power of deep learning systems, but rather to shed light on the current trend in literature, where classical methods are not carefully considered for comparison although they can become powerful baselines if designed properly.

{\small
\bibliographystyle{unsrt}
\bibliography{depth_comp_bib}
}

\end{document}